\documentclass{article}
\usepackage{spconf,amsmath,graphicx,hyperref,marvosym,multirow,xcolor,colortbl,booktabs,amsfonts}
\usepackage[normalem]{ulem}
\useunder{\uline}{\ul}{}

\newcommand{\ModelName}{Think-Clip-Sample}
\newcommand{\ModelAbbr}{TCS}
\definecolor{mygreen}{HTML}{00AA00}
\title{Think-Clip-Sample: Slow-Fast Frame Selection for Video Understanding}
\twoauthors
 {Wenhui Tan\sthanks{Work performed when Wenhui Tan was a research intern at Xiaomi Inc.}, Ruihua Song\textsuperscript{\Letter}}
	{Gaoling School of Artificial Intelligence\\
	Renmin University of China\\
	Beijing, China}
 {Jiaze Li, Jianzhong Ju, Zhenbo Luo\textsuperscript{\Letter}}
	{MiLM Plus\\
	Xiaomi Inc.\\
	Beijing, China}

\begin{document}

\ninept
\maketitle
\begin{abstract}
Recent progress in multi-modal large language models (MLLMs) has significantly advanced video understanding. However, their performance on long-form videos remains limited by computational constraints and suboptimal frame selection. We present Think-Clip-Sample (TCS), a training-free framework that enhances long video understanding through two key components: (i) \textbf{Multi-Query Reasoning}, which generates multiple queries to capture complementary aspects of the question and video; and (ii) \textbf{Clip-level Slow-Fast Sampling}, which adaptively balances dense local details and sparse global context. Extensive experiments on MLVU, LongVideoBench, and VideoMME demonstrate that TCS consistently improves performance across different MLLMs, boosting up to 6.9\% accuracy, and is capable of achieving comparable accuracy with 50\% fewer inference time cost, highlighting both efficiency and efficacy of TCS on long video understanding.

\end{abstract}

\renewcommand{\thefootnote}{} 
\footnotetext{\Letter~Corresponding authors: Ruihua Song and Zhenbo Luo.}

\section{Introduction}\label{sec:intro}

Understanding long-form videos has become a central challenge in advancing multi-modal large language models (MLLMs)~\cite{gpt4o,seed15vl,imove}.
A long video often consists of thousands of frames, even with a low frame-per-second (FPS) sampling rate, leading to prohibitive computational costs.
Mainstream MLLMs typically adopt uniform frame sampling~\cite{qwen2vl,qwen25vl,internvl}, but this strategy treats all frames equally, regardless of their informativeness, thus resulting in subpar performance.

To mitigate this issue, several methods have been proposed to select more informative frames, e.g., Q-Frame~\cite{qframe} introduces adaptive frame selection and multi-resolution scaling tailored to the video content and the specific query. It proposes a CLIP-based~\cite{clip} similarity with the Gumbel-Max trick for efficient selection, allowing Video-LLMs to process more relevant frames without exceeding computational limits. 
AKS~\cite{aks} formulates selection as an optimization problem balancing query relevance and video coverage, and provides an adaptive algorithm to approximate the optimal solution.

Despite these advances, existing approaches face two fundamental challenges. First, they rely solely on direct question-frame similarity, assuming the question adequately represents all information needs. However, questions are often abstract and incomplete. For instance, a question like ``\textit{Which team won in the end? (A) Team in black clothes; (B) Team in white clothes}'' only mentions subjects while omitting crucial details about actions and context. Directly feeding such questions to CLIP often retrieves frames showing players without capturing game-decisive moments.
Second, similarity-based sampling tends to produce unbalanced frame distributions. Algorithms may duplicate spikes of high similarity frames, while neglecting informative regions with moderate similarity scores and global context. This results in sparse coverage that misses important information for comprehensive video understanding.

In this paper, we propose \textbf{\ModelName~(\ModelAbbr)}, a training-free method designed to address these challenges. 
(1) To enhance diversity and ensure broader coverage,~\ModelAbbr~first \textbf{thinks} of multiple queries~\cite{o1,ttr,r1,colar}.
Instead of relying on a single question, we prompt the MLLM to generate multiple queries from different perspectives (e.g., objects, scenes, actions). These multi-view queries encourage the MLLM to capture complementary information for answering the question across diverse video segments.
(2) To avoid sparse and uneven sampling, we propose a Clip-level Slow-Fast Sampling strategy: Given a total frame budget $K$, \ModelAbbr~first identify high-similarity \textbf{clips} rather than isolated frames. Then, a larger portion of $K$ is allocated to these compact yet informative clips, denoted as slow \textbf{sampling}, while the remaining frames are uniformly sampled from non-clipped regions (fast sampling) to provide global coverage. This balanced allocation ensures both fine-grained detail and a global context.

We evaluate \ModelAbbr~on two base MLLMs, Qwen2-VL-7B~\cite{qwen2vl} and MiMo-VL-7B~\cite{mimo}, across three challenging benchmarks: LongVideo-Bench~\cite{lvbench}, MLVU~\cite{mlvu}, and VideoMME~\cite{videomme}. Experimental results demonstrate that \ModelAbbr~consistently improves performance of the base MLLMs under the same sampling budget, achieving up to 6.9\% accuracy improvement on MLVU, while reducing computation time by over 50\% on Qwen2-VL-7B with comparable performance, highliting both efficiency and efficacy of \ModelAbbr~for long video understanding.

\begin{figure*}[t]
    \centering
    \includegraphics[width=\linewidth]{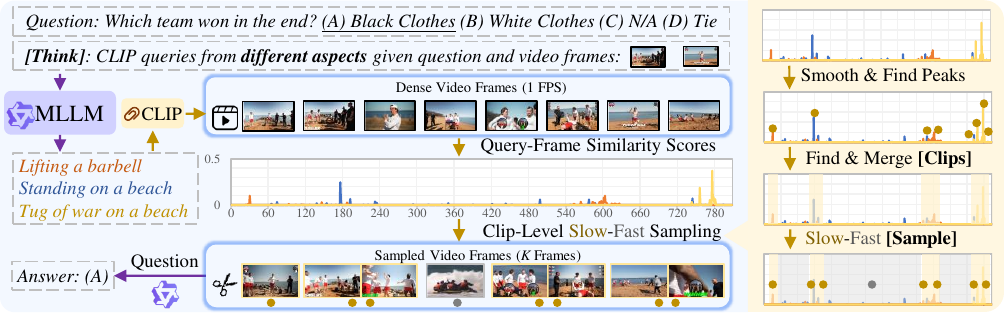}
    \caption{Overview of our proposed method~\ModelName~(\ModelAbbr): (i) Given a question on a long video, \ModelAbbr~first \textbf{thinks} of queries from different perspectives, which are used with CLIP to retrieve frames with both high relevance and broad coverage. (ii) Instead of sampling frames with highest similarity scores,~\ModelAbbr~identifies high-relevance~\textbf{clips}, and then (iii) applies Slow-Fast \textbf{Sampling} to allocate more frames on informative clips (yellow) and distribute the remainder across non-clip regions (gray), preserving both local detail and global context.}
    \label{fig:method}
\end{figure*}

Our main contributions are threefold:
\begin{itemize}
    \item We propose \textbf{Multi-Query Reasoning}, to automatically generate multiple queries from the question and video, enabling the retrieval of diverse and complementary frames from different perspectives such as objects, actions, and scenes.
    \item We introduce \textbf{Clip-level slow-fast sampling} strategy, which first identifies high-similarity clips and allocates a larger portion of the frame budget to these regions, and uniformly sampling the remainder from non-clipped regions, to ensure both fine-grained detail and global coverage.
    \item Experimental results demonstrate that \ModelAbbr~significantly improves long video understanding across multiple MLLMs and benchmarks, achieving up to 6.9\% accuracy gain, and reducing inference time by over 50\% with comparable performance, demonstrating both efficiency and efficacy.
\end{itemize}

\section{Method}\label{sec:method}

\subsection{Overview}\label{sec:method:overview}
This work focuses on video understanding through video question answering tasks. Given a video $V \in \mathbb{R}^{T \times H \times W \times 3}$ with $T$ frames and $(H \times W)$ resolution, and a multi-option question $Q$, the goal of the MLLM is to select the correct answer among the options.

For long videos (e.g., longer than 10 minutes), $T$ can be very large and infeasible to process directly. Thus, selecting a subset of $K$ frames, where $K \ll T$, is crucial for efficient video understanding. A straightforward approach is to uniformly sample $K$ frames across the video. Although this ensures temporal coverage, it often results in sparse observations that miss crucial visual evidence.

To address this, a common strategy is to compute similarity scores $\mathbf{s} = \{s_i\}_{i=1}^T$ between the question and frames using a pre-trained vision-language models (VLMs)~\cite{xvlm,blip,siglip,llavasg,li2025multilevel}, such as CLIP~\cite{clip}, and then sample frames according to $\mathbf{s}$. 
However, this pipeline presents two key challenges: (1) how to obtain better similarity scores $\mathbf{s}$, and (2) how to design an effective sampling strategy given $\mathbf{s}$. We address these in Section~\ref{sec:method:query} and~\ref{sec:method:sample}, respectively.

\subsection{Reasoning for Multi-Perspective Queries}\label{sec:method:query}
Previous work typically computes similarity scores directly between the \textbf{question} and video frames $V$ with VLMs such as CLIP. However, these VLMs are designed to compare visual descriptions with images, rather than natural language questions. Furthermore, a single question may only highlight one aspect of the visual content, resulting in incomplete frame retrieval. For example, given a question such as ``\textit{Which team won in the end? (A) Team in Black Clothes (B) Team in White Clothes}’’, CLIP would retrieve all the frames where the players are present, lacking core relevance to the question.

To overcome these limitations, we propose \textbf{Multi-Query Reasoning}. As shown in Figure~\ref{fig:method} (left), instead of feeding the question directly into the CLIP model, we first provide the question to the MLLM with a small set of sparsely sampled low-resolution frames to provide neccesary semantics.
Then, the MLLM is prompted to generate multiple queries $\mathbf{q} = \{q_i\}_{i=1}^{N_q}$, each from a different perspective, such as objects, scenes, or actions potentially relevant to the question. To balance coverage and efficiency, we limit the maximum number of queries to $N_q = 4$. Each query is then passed to CLIP to compute similarity scores $\mathbf{s}_{mq} \in \mathbb{R}^{N_q \times T}$ with frames sampled in 1 FPS. Finally, we aggregate across queries using average pooling to obtain $\mathbf{s}$, yielding frame-level similarity scores enriched by multi-perspective evidence.

\subsection{Clip-level Slow-Fast Sampling}\label{sec:method:sample}
With the similarity scores $\mathbf{s}$, a naive method is top-k sampling, which selects frames with the highest scores.
However, as $\mathbf{s}$ is aggregated from multiple perspectives, it naturally follows a \textbf{multimodal distribution}. Using top-k sampling may overly focus on sharp score spikes, neglecting other informative regions and global observation.

To address this, we propose a \textbf{Clip-level Slow-Fast Sampling} strategy, as illustrated in Figure~\ref{fig:method} (right). The core idea is to identify short but informative clips and allocate a majority of frames (the \textbf{slow} path) within these clips, while dedicating a smaller portion of frames (the \textbf{fast} path) to uniformly sample the remaining video, thus maintaining global context.

Concretely, we first obtain smoothed similarity scores $\mathbf{s}$ with a Gaussian filter to reduce noise:
\begin{equation}
    \mathbf{s}_{\text{smoothed}}[i] = \frac{1}{\sqrt{2\pi}\sigma} \sum_{j=-r}^{r} \mathbf{s}[i+j] \cdot \exp\left(-\frac{j^2}{2\sigma^2}\right),
\end{equation}
where we use default values of kernel radius $r=4$ and $\sigma=1$.

We then compute a dynamic threshold $\tau_s = \mu_s + \alpha \sigma_s$, where $\mu_s$ and $\sigma_s$ are the mean and standard deviation of $\mathbf{s}_{\text{smoothed}}$, and $\alpha$ is a hyperparamter. Local maxima above $\tau_s$ are detected as peaks, and we expand around each peak with decreasing scores to form candidate clips. Finally, we merge overlapping clips to avoid duplication.

\begin{table*}[t]
\centering
\caption{Comparison of MLLMs and baseline methods against our method~\ModelAbbr~on three benchmarks. We bold the best and underline the second best results. The performance of baseline methods are reported according to original papers.}
\label{tab:main}
\begin{tabular}{l|ccccccc}
\toprule
            & \multirow{2}{*}{\#Frames} & \multirow{2}{*}{MLVU} & \multirow{2}{*}{LongVideoBench} & \multicolumn{4}{c}{VideoMME (w/o sub.)}                       \\
            &                           &                       &                                 & Overall       & Short         & Medium        & Long          \\ \midrule
\textbf{\textit{Long-form Video-LLMs}}\\
Video-XL-7B \textit{(CVPR 2025)} & 128                         & {\ul 64.9}                  & -                            & 55.5          & 64.0          & 53.2          & \textbf{49.2}          \\
LongVILA-8B \textit{(ICLR 2025)} & 128                        & -                     & -                            & 49.2          & 60.2             & 48.2             & 38.8             \\ \midrule
\textbf{\textit{Frame-Sampling Methods}}\\
Qwen2-VL-7B & 32                        & 58.1                     & 55.5                            & 57.6          & -             & -             & -             \\
w/ AKS \textit{(CVPR 2025)}         & 32                        & -                     & {\ul 60.5}                      & \textbf{59.9} & -             & -             & -             \\
w/ Q-Frame \textit{(ICCV 2025)}     & 4+8+32                    & \textbf{65.4}         & 58.4                            & {\ul 58.3}    & \textbf{69.4} & 57.1          & {\ul 48.3} \\
\textbf{w/ TCS} (\textit{Ours})         & 32                        & 61.2            & \textbf{60.9}                   & 58.0          & {\ul 69.2}    & \textbf{57.4} & 47.9    \\ \midrule
MiMo-VL-7B  & 32                        & 60.9                  & 64.3                            & 62.4          & 75.2          & 60.7          & 51.3          \\
\textbf{w/ TCS} (\textit{Ours})         & 32                        & \textbf{67.8}~(6.9$\uparrow$)         & \textbf{67.2}~(2.9$\uparrow$)                   & \textbf{65.0}~(2.6$\uparrow$) & \textbf{76.3}~(1.1$\uparrow$) & \textbf{65.9}~(5.2$\uparrow$) & \textbf{52.7}~(1.4$\uparrow$) \\ \bottomrule
\end{tabular}
\end{table*}

Then we devide the budget of $K$ frames into $K_{\text{slow}}$ and $K_{\text{fast}}$ (e.g., $K_{\text{slow}} = 3K/4$, $K_{\text{fast}} = K/4$), and proceed as follows:
\begin{itemize}
    \item \textbf{Slow sampling.} We uniformly sample $K_{\text{slow}}$ frames from all frames within the detected clips (in yellow). This ensures dense and uniform coverage of locally informative regions. If the total number of clip frames is smaller than $K_{\text{slow}}$, indicating too short high-relevance clips, we divide the peak threshold $\alpha$ by 2 to enlarge the clips, and re-process the clipping.
    \item \textbf{Fast sampling.} We uniformly sample $K_{\text{fast}}$ frames from the remaining non-clipped regions (in gray). This guarantees that frames outside high-similarity regions still contribute to the overall context. If the non-clip regions contain fewer than $K_{\text{fast}}$ frames, denoting too many high-relevance clips, we multiply the peak threshold $\alpha$ by 2 and re-clip.
\end{itemize}

Finally, we merge the slow and fast sampled frames to construct the final set of $K$ frames. This hybrid allocation ensures that key segments are well-represented without sacrificing global video coverage, balancing local detail and global understanding.

\section{Experiments}\label{sec:exp}

\subsection{Experimental Setup}\label{sec:exp:setup}
\textbf{Baseline Methods.}
To evaluate the effectiveness and efficiency of~\ModelAbbr, we implement and compare our method on two base MLLMs: \textbf{Qwen2-VL-7B}~\cite{qwen2vl} and \textbf{MiMo-VL-7B}~\cite{mimo}.
We also include two long-form Video-LLMs and two training-free frame selection methods:
(i) \textbf{Video-XL}~\cite{videoxl} leverages MLLMs’ inherent key-value (KV) sparsification capacity to condense the visual input;
(ii) \textbf{LongVILA}~\cite{longvila} introduces long-context Multi-Modal Sequence Parallelism for high efficiency long video input training;
(iii) \textbf{AKS}~\cite{aks} adopts adaptive keyframe selection by detecting saliency peaks across temporal similarity distributions;  
(iv) \textbf{Q-Frame}~\cite{qframe} integrates semantic relevance into dynamic-resolution frame selection.\\
\textbf{Benchmarks.}
We evaluate on three widely used long video understanding benchmarks:  
(i) \textbf{MLVU}~\cite{mlvu}, consisting of 2,593 questions across nine categories, with average video duration of 12 minutes.  
(ii) \textbf{LongVideoBench}~\cite{lvbench}, a benchmark targeting long video comprehension, with 1,337 questions and similar average video duration.  
(iii) \textbf{VideoMME}~\cite{videomme}, which covers Short (1.3 min), Medium (9 min), and Long (41 min) subsets, each with 900 questions from 300 videos.  
To focus on pure visual understanding, we run all benchmarks without subtitles.\\
\textbf{Implementation Details.} For Multi-Query Reasoning, we uniformly sample $K/4$ frames with minimum resolution (up to $224 \times 224$) for lightweight visual prompts. CLIP-ViT-Large-FP16~\cite{clip} is adopted as the query-frame similarity scorer.  
For Clip-level Slow-Fast Sampling, the peak threshold $\alpha$ is set to 0.5 and the ratio of fast frames is set to 1/4 across all the benchmarks.
All experiments are conducted on vLLM backend~\cite{vllm}.

\subsection{Main Results}\label{sec:exp:main}
Table~\ref{tab:main} reports the performance of \ModelAbbr~against baseline methods on both Qwen2-VL-7B and MiMo-VL-7B.  
Several key observations can be made:\\
(1) Compared to long-form Video-LLMs, frame-sampling based methods achieves higher average performance with fewer input frames, highlighting the efficiency and efficacy of key frame sampling in long video understanding.\\
(2) Compared to AKS and Q-Frame,~\ModelAbbr~achieves the best performance on LVBench and VideoMME-Medium. We mainly attribute this to our multi-query design, which provides comprehensive cues to answer the question, along with our clip-level slow-fast sampling strategy that captures both local and global information.\\
(3) \ModelAbbr~consistently improves performance of the two base MLLMs across all the benchmarks: Qwen2-VL-7B and MiMo-VL-7B gain an average of 3.0\% and 4.1\% accuracy improvement, respectively. These gains validate the effectiveness and generalizability of the Multi-Query and Slow-Fast designs.
Moreover, we notice that \ModelAbbr~expresses more significant performance gain with MiMo-VL-7B (6.9\% on MLVU, 2.9\% on LVBench, and 5.2\% on VideoMME-Medium). This could be attribute to the fact that MiMo-VL is fundamentally a ``\textit{reasoning model}''~\cite{o1,r1}, i.e., it was originally trained on complex reasoning tasks, which enables more efficient multi-query reasoning, producing more comprehensive queries that better leverage our framework's capabilities.

\begin{figure*}[t]
    \centering
    \includegraphics[width=\linewidth]{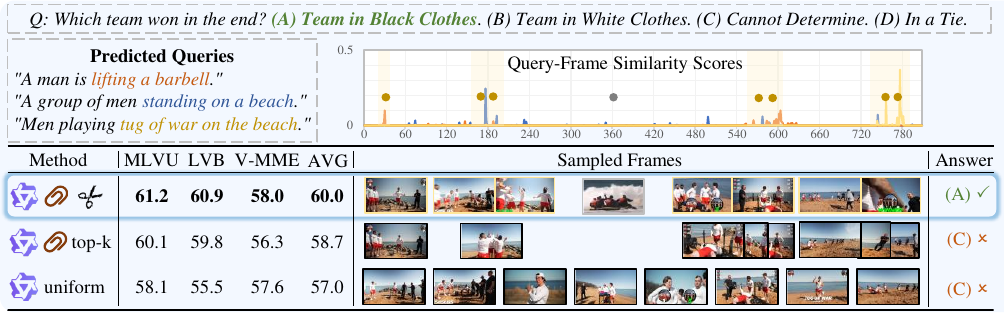}
    \caption{A case study on VideoMME and ablation results on Multi-Query Reasoning and Clip-level Slow-Fast Sampling component.}
    \label{fig:case}
\end{figure*}

\subsection{Qualitative and Quantative Ablation Study}\label{sec:exp:abla}
In this section, we answer the question: ``\textit{How does our proposed components influence model performance qualitatively and quantitively?}''.
The experiments are conducted on Qwen2-VL-7B and the results are shown with Figure~\ref{fig:case}.

\textbf{Quantitatively}, comparing line 2 vs. line 3 shows that Multi-Query Reasoning boosts accuracy by an average of 1.7\%. Comparing line 2 vs. line 1 shows that Slow-Fast Sampling further improves accuracy by 1.3\%. Together, these results confirm that both components contribute complementary benefits. 

\textbf{Qualitatively}, for a question such as \textit{``which team won in the end''}, Multi-Query Reasoning successfully identifies key event queries (\textit{lifting}, \textit{standing on the beach}, \textit{tug-of-war}), allowing CLIP to retrieve highly relevant frames. Without it, retrieval would degenerate to semantically sparse frames (\textit{people in black/white clothes}).  
In the second stage, Slow-Fast Sampling captures critical sequences (e.g., lifting, bench press, arm wrestling, tug-of-war), including a decisive frame with “winning” in the background. In contrast, top-k focuses too narrowly on dense \textbf{redundant} clips, missing long-range evidence, while uniform sampling provides broad but \textbf{uninformative} coverage. This highlights~\ModelAbbr’s advantage in balancing fine-grained evidence with global temporal reasoning.

\begin{figure}
    \centering
    \includegraphics[width=\linewidth]{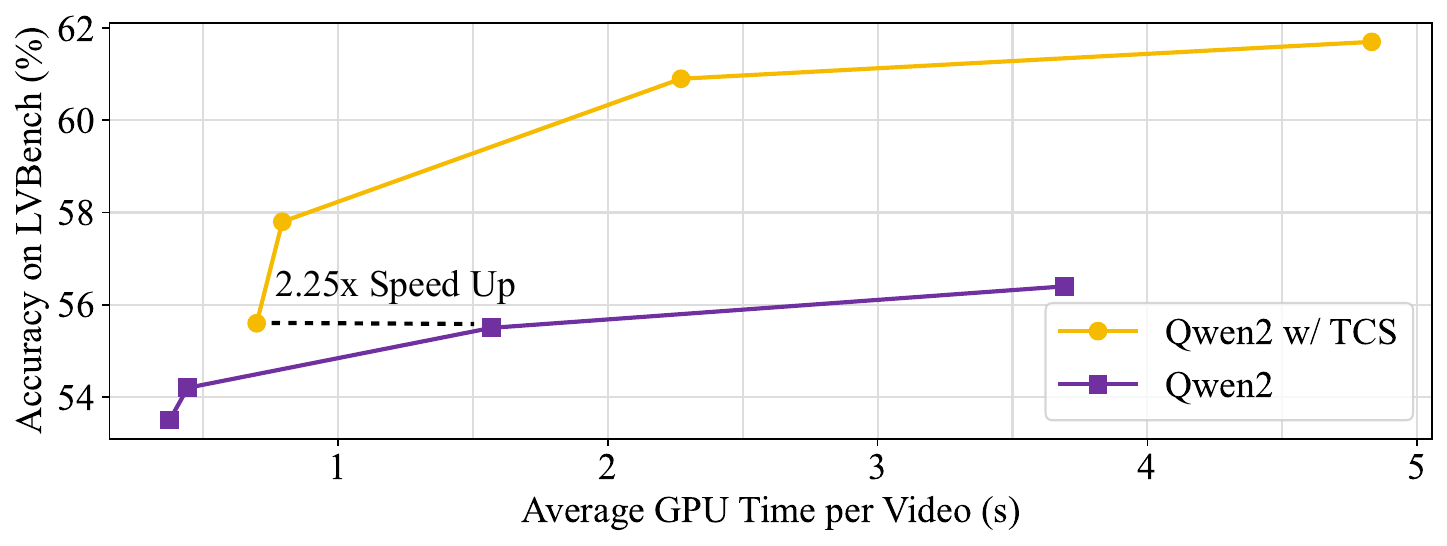}
    \caption{Speed-accuracy curves of \ModelAbbr~and Qwen2-VL-7B.}
    \label{fig:speed}
\end{figure}

\subsection{Efficiency and Parameter Sensitivity Study}\label{sec:exp:para}
In this section, we answer the following question: ``\textit{How efficient is \ModelAbbr~and how does} $K_{\text{fast}}/K$ \textit{and $\alpha$ influence performance?}''. The experiments are performed on LongVideoBench and Qwen2-VL-7B.

\textbf{To assess efficiency}, we evaluate~\ModelAbbr~against Qwen2-VL-7B with varying input budgets $K\in\{8,16,32,64\}$ (Figure~\ref{fig:speed}). Although query generation and CLIP inference introduce slight overhead,~\ModelAbbr~consistently yields higher accuracy under all budgets.
Crucially, for a target accuracy of $\sim$55\%, the base model requires 32 frames and $\sim$1.5s inference, while \ModelAbbr~achieves this with only 8 frames and $\sim$0.7s, yielding over $2\times$ speedup. This demonstrates that~\ModelAbbr~not only improves accuracy but also substantially enhances efficiency in real-world long video applications.
We have to mention that: thanks to vLLM's~\cite{vllm} excellent inference technique, the question-answering process could be significantly accelerated with similar KV-cache stored in the query reasoning process.

\begin{figure}
    \centering
    \includegraphics[width=\linewidth]{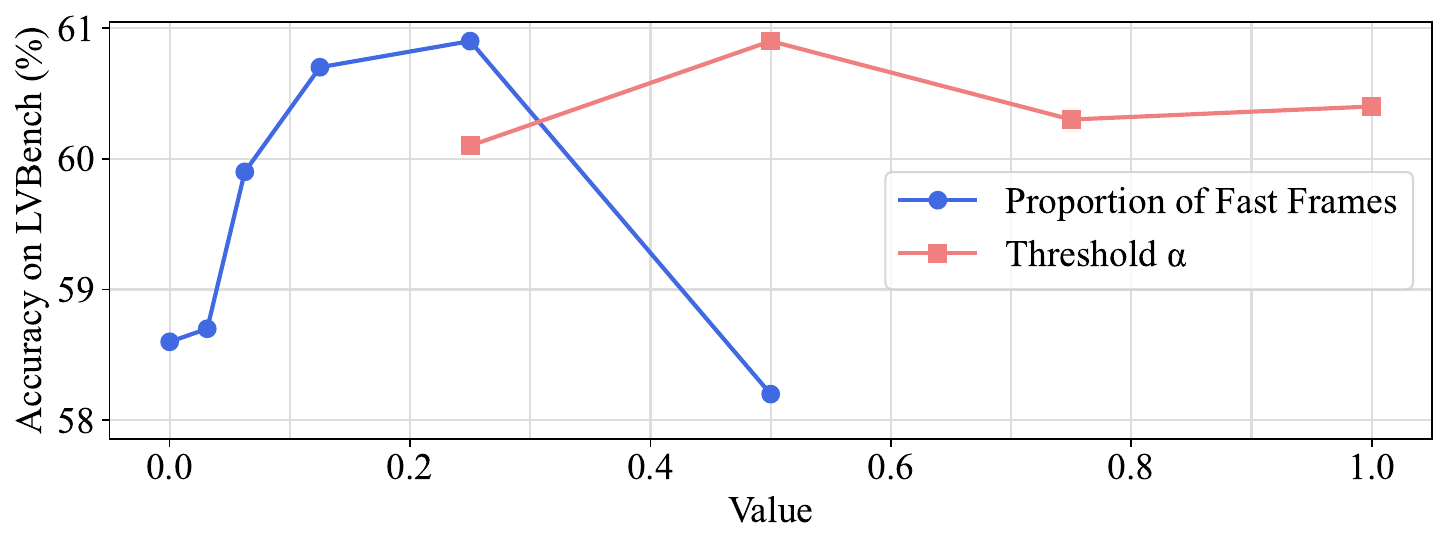}
    \caption{Parameter analyses on fast frame proportion and threshold $\alpha$.}
    \label{fig:parasense}
\end{figure}

To evaluate the robustness of our method, we conduct experiments by varying two key hyperparameters: the proportion of fast frames in the Slow-Fast Sampling stage and the threshold $\alpha$ for locating high-similarity peaks.
Specifically, we test fast proportions $K_{\text{fast}}/K \in \{1/32, 1/16, 1/8, 1/4, 1/2\}$, and threshold values $\alpha \in \{1/4, 1/2, 3/4, 1\}$:\\
\textbf{Fast frame proportion.} As shown in Figure~\ref{fig:parasense}, when the proportion of fast frames is set too low (e.g., $0$ or $1/32$), the overall performance drops to about $58.5\%$, mainly due to insufficient coverage of global video context. Conversely, when the fast proportion is too high (e.g., $1/2$), the accuracy decreases to around $58.0\%$, since excessive fast frames reduces the ability to capture fine-grained details within regions of interest and introduces background noise. We observe that allocating approximately $1/4$ of frames to the fast pathway achieves the best balance, providing both local detail and global context.\\
\textbf{Peak threshold $\alpha$.} A similar trend is observed when varying $\alpha$. Setting $\alpha$ too low (e.g., $1/4$) produces an excessive number of clips, which dilutes the $K_{\text{slow}}$ budget and reduces accuracy to about $60\%$. On the other hand, setting $\alpha$ too high (e.g., $1$) results in only a few detected clips, leading to the omission of potential regions of interest and a similar performance degradation. The best results are obtained when $\alpha$ is set to moderate values around $1/2$, striking a balance between coverage and precision.

\section{Conclusion}\label{sec:conclusion}
In this paper, we introduce \ModelName~(\ModelAbbr), a training-free frame-selection framework for MLLMs long video understanding with two key innovations: \textbf{Multi-Query Reasoning} for comprehensive visual information capture, and \textbf{Clip-level Slow-Fast Sampling} for balanced frame allocation. Experiments on three benchmarks show \ModelAbbr~achieves up to 6.9\% accuracy gains over backbone MLLM, and is capable of reducing inference time by 50\% with comparable performance, making it valuable for resource-constrained applications. Future work will explore incorporating audio and speech signals for complex video understanding.

\vfill\pagebreak




\bibliographystyle{IEEEbib}
\bibliography{refs}

\end{document}